\documentclass[10pt,twocolumn,letterpaper]{article}

\usepackage{graphicx}
\usepackage{subcaption}
\usepackage{float}
\usepackage[justification=raggedright]{caption}	
\usepackage{lscape}                                         
\usepackage{caption}
\usepackage[percent]{overpic}

\usepackage[lined,ruled,linesnumbered]{algorithm2e}

\usepackage{booktabs}                   
\usepackage{multirow}

\usepackage{paralist}
\usepackage{enumitem}

\usepackage{bm}                          
\usepackage{epsfig}                      
\usepackage{graphicx}                  
\usepackage{times}
\usepackage{mathtools}
\usepackage{amssymb,amsmath}   

\usepackage{units}
\usepackage{color}

\usepackage{comment}

\usepackage{url}  
\usepackage[pagebackref=true,breaklinks=true,colorlinks,bookmarks=false,citecolor=blue,linkcolor=blue]{hyperref}

\usepackage{xspace}
\usepackage[table]{xcolor}
\usepackage{setspace}
\usepackage{gensymb}

\def\etal{et~al.\_}			  
\def\eg{e.g.,~}               
\def\ie{i.e.,~}               

\def\naive{na{\"i}ve\xspace}

\DeclareMathOperator*{\argmin}{\arg\!\min}

\newlength\paramarginsize
\newlength\figmarginsize
\newlength\secmarginsize
\newlength\figcapmarginsize
\newlength\tabcapmarginsize

\newcommand{\paramargin}{\vspace{\paramarginsize}}

\newcommand{\mpage}[2]
{
\begin{minipage}{#1\linewidth}\centering
#2
\end{minipage}
}

\newcommand{\topic}[1]
{\vspace{1mm}\paramargin\noindent\textbf{#1}}

\newcommand{\heading}[1]
{\vspace{1mm}\noindent\textbf{#1}}

\newcommand{\secref}[1]{Section~\ref{sec:#1}}
\newcommand{\figref}[1]{Figure~\ref{fig:#1}} 
\newcommand{\tabref}[1]{Table~\ref{tab:#1}}

\long\def\ignorethis#1{}
\definecolor{aqua}{rgb}{0.1, 0.5, 0.7}
\definecolor{tealblue}{rgb}{0.212, 0.459, 0.533}

\newcommand{\tb}[1]{\textbf{#1}}

\def\xi{\mathbf{x}_i}

\graphicspath{{figure}, {images}, {example}}

\usepackage{iccv} 

\iccvfinalcopy 

\def\iccvPaperID{2875} 

\ificcvfinal\fi
\begin{document}

\title{Portrait Neural Radiance Fields from a Single Image}
\author{
Chen Gao\\
Virginia Tech\\
\and
Yichang Shih\\
Google\\
\and
Wei-Sheng Lai\\
Google\\
\and
Chia-Kai Liang\\
Google\\
\and
Jia-Bin Huang\\
Virginia Tech\\
}

\twocolumn[{
\renewcommand\twocolumn[1][]{#1}
\maketitle
\centering \vspace{-12mm}\url{https://portrait-nerf.github.io} 
\newlength\figwidthA
\setlength\figwidthA{0.195\linewidth}

\newlength\figwidthB
\setlength\figwidthB{0.39\linewidth}

\begin{center}
\centering
\footnotesize

\newcommand{\teaser}[1]{\includegraphics[trim=0cm 0cm 0cm 2cm, clip=true, width=\figwidthA]{fig/Teaser/#1.png}\hfill}
\newcommand{\teaserfovb}[1]{\includegraphics[trim=2.5cm 4.4cm 2.5cm 3.5cm, clip=true, width=\figwidthA]{fig/Teaser/#1.png}\hfill}

\newcommand{\teaserfova}[1]{\includegraphics[trim=2.4cm 4.1cm 2.2cm 3.3cm, clip=true, width=\figwidthA]{fig/Teaser/#1.png}\hfill}

\teaser{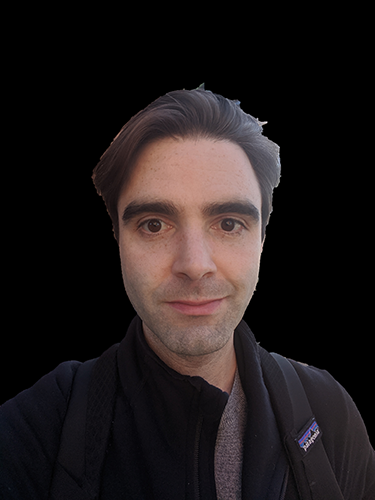}
\teaser{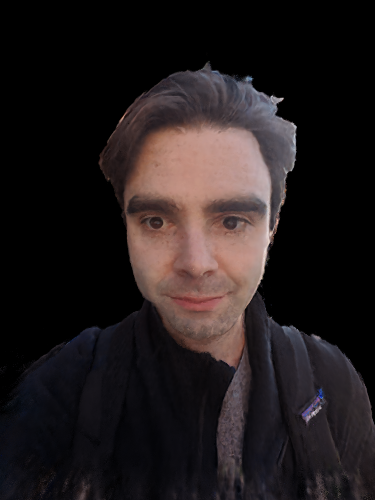}
\teaser{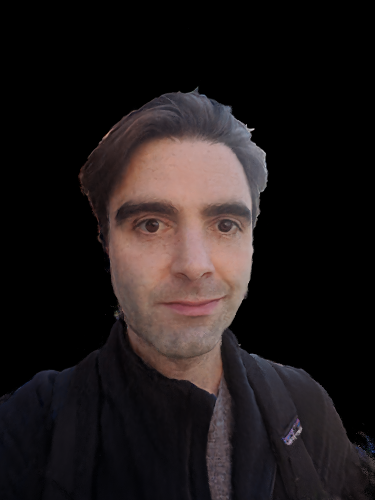}
\teaserfovb{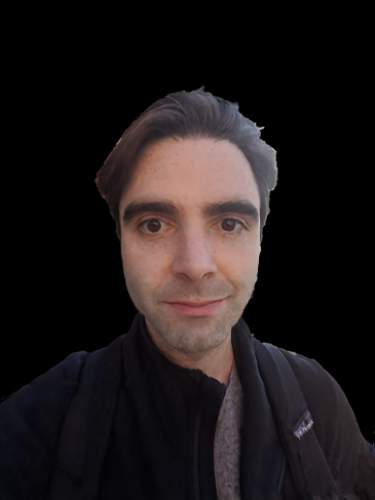}
\teaserfova{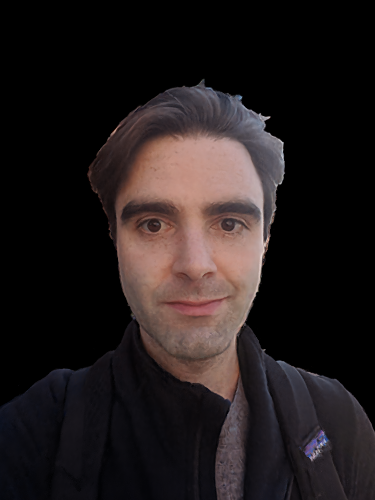}


\vspace{-3mm}
$\underbracket[1pt][2.0mm]{\hspace{\figwidthA}}_%
    {\substack{\vspace{-2.0mm}\\\colorbox{white}
    {(a) Input}}}$
\vspace{2mm} \hfill
$\underbracket[1pt][2.0mm]{\hspace{\figwidthB}}_%
    {\substack{\vspace{-2.0mm}\\\colorbox{white}
    {(b) Novel view synthesis}}}$\vspace{2mm} \hfill
$\underbracket[1pt][2.0mm]{\hspace{\figwidthB}}_%
    {\substack{\vspace{-2.0mm}\\\colorbox{white}
    {(c) FOV manipulation}}}$\vspace{2mm}

\vspace{-5mm}
\captionof{figure}{
\textbf{Applications of the proposed method.}
Given a single portrait image (a) as input, our method produces a \emph{portrait neural radiance field} that facilitates photorealistic face editing tasks such as (b) novel view synthesis and (c) field-of-view (FOV) manipulation, where the left and right are rendered with wider and narrower camera FOVs, respectively. 
}
\label{fig:teaser}
\end{center}
}]

\begin{abstract}
We present a method for estimating Neural Radiance Fields (NeRF) from a single headshot portrait.
While NeRF has demonstrated high-quality view synthesis, it requires multiple images of static scenes and thus impractical for casual captures and moving subjects.
In this work, we propose to pretrain the weights of a multilayer perceptron (MLP), which implicitly models the volumetric density and colors, with a meta-learning framework using a light stage portrait dataset.
To improve the generalization to unseen faces, we train the MLP in the canonical coordinate space approximated by 3D face morphable models.
We quantitatively evaluate the method using controlled captures and demonstrate the generalization to real portrait images, showing favorable results against state-of-the-arts.

\vspace{-5mm}
\end{abstract}

\section{Introduction}
\label{sec:intro}

Portrait view synthesis enables various post-capture edits and computer vision applications, such as pose manipulation~\cite{Criminisi-2003-GMF}, selfie perspective distortion (foreshortening) correction~\cite{Zhao-2019-LPU,Fried-2016-PAM,Nagano-2019-DFN}, improving face recognition accuracy by view normalization~\cite{Zhu-2015-HFP}, and greatly enhancing the 3D viewing experiences. 
Compared to 3D reconstruction and view synthesis for generic scenes, portrait view synthesis holds a higher quality standard to avoid the uncanny valley.

To achieve high-quality view synthesis, the filmmaking production industry densely samples lighting conditions and camera poses using a light stage~\cite{Debevec-2000-ATR}, which requires expensive hardware setups unsuitable for casual users. 
Reconstructing the facial geometry from a \emph{single capture} requires face mesh templates~\cite{Bouaziz-2013-OMF} or a 3D morphable model~\cite{Blanz-1999-AMM,Cao-2013-FA3,Booth-2016-A3M,Li-2017-LAM}.
While the quality of these 3D model-based methods has been improved dramatically via deep networks~\cite{Genova-2018-UTF,Xu-2020-D3P}, a common limitation is that the model only covers the center of the face and excludes the upper head, hairs, and torso. 
We emphasize that the excluded regions are critical for natural portrait view synthesis.

Recently, neural implicit representations emerge as a promising way to model the appearance and geometry of 3D scenes and objects \cite{sitzmann2019scene,Mildenhall-2020-NRS,liu2020neural}.
For example, Neural Radiance Fields (NeRF) demonstrates high-quality view synthesis by implicitly modeling the volumetric density and color using the weights of a multilayer perceptron (MLP).
However, training the MLP requires capturing images of \emph{static subjects} from \emph{multiple viewpoints} (in the order of 10-100 images)~\cite{Mildenhall-2020-NRS,Martin-2020-NIT}.
It is thus impractical for portrait view synthesis because a slight subject movement or inaccurate camera pose estimation degrades the reconstruction quality.


In this paper, we propose to train an MLP for modeling the radiance field using a \emph{single headshot portrait} illustrated in Figure~\ref{fig:teaser}.
Unlike NeRF~\cite{Mildenhall-2020-NRS}, training the MLP with a single image from scratch is fundamentally ill-posed, because there are infinite solutions where the renderings match the input image.
Our key idea is to pretrain the MLP and finetune it using the available input image, and adapt the model to an unseen subject's appearance and shape.
To pretrain the MLP, we use densely sampled portrait images in a light stage capture.
However, using a \naive pretraining process that optimizes the reconstruction error between the synthesized views (using the MLP) and the rendering (using the light stage data) over the subjects in the dataset performs poorly for \emph{unseen} subjects due to the diverse appearance and shape variations among humans.
We address the challenges in two novel ways.
First, we leverage gradient-based meta-learning techniques~\cite{Finn-2017-MAM} to train the MLP that quickly adapts to an unseen subject. 
Second, we propose to train the MLP in a canonical coordinate by exploiting domain-specific knowledge about the face shape.
Our experiments show favorable quantitative results against the state-of-the-art 3D face reconstruction and synthesis as well as 2D GAN-based algorithms on the dataset of controlled captures. 
We show $+5$dB PSNR improvement against the second best approach in ground truth validation.
We further show that our method performs well for real input images captured in the wild and demonstrate foreshortening distortion correction as an application. 

In this work, we make the following contributions:
\begin{compactitem}
\item 
We present a single-image view synthesis algorithm for portrait photos by leveraging meta-learning. Our method produces a \emph{full reconstruction}, covering not only the facial area but also the upper head, hairs, torso, and accessories such as eyeglasses. 
\item 
We propose an algorithm to pretrain NeRF in a canonical face space using a rigid transform from the world coordinate. 
We show that compensating the shape variations among the training data substantially improves the model generalization to unseen subjects. 
\item 
We provide a multi-view portrait dataset consisting of controlled captures in a light stage. 
Our data provide a way of quantitatively evaluating portrait view synthesis algorithms.
\end{compactitem}

\section{Related Work}
\label{sec:related}

\begin{figure*}[t]
\footnotesize
\begin{overpic}[width=1\linewidth]{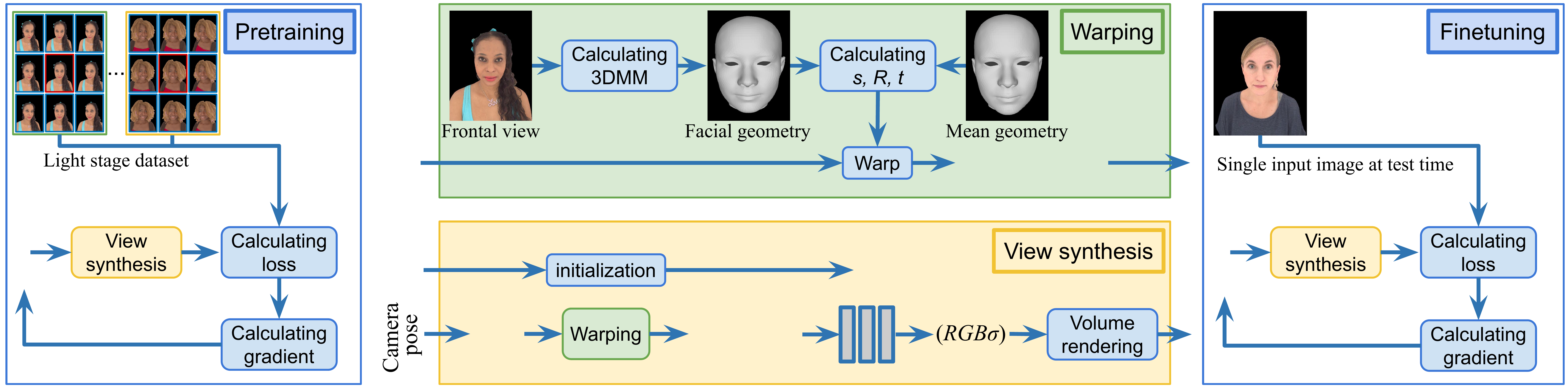}
\put (30,3) {$ {\scriptstyle (\vec{x}, \vec{d})} $}
\put (44,3) {$ {\scriptstyle (\mathit{s}\mathbf{R}\vec{x}+\mathbf{t}, \vec{d})} $}
\put (62,13.7) {$ {\scriptstyle (\mathit{s}\mathbf{R}\vec{x}+\mathbf{t}, \vec{d})} $}
\put (25.5,7) {$ {\scriptstyle \theta} $}
\put (54.5,7) {$ {\scriptstyle f_\theta} $}
\put (25,14) {\rotatebox[origin=c]{90}{$ {\scriptstyle (\vec{x}, \vec{d})} $}}
\put (0.5,7.7) {$ {\scriptstyle \theta_{p,m}} $}


\put (77,7.7) {$ {\scriptstyle \theta_{s}} $}
\end{overpic}

\hspace{10mm}{(a) Pretrain NeRF}
\hspace{40mm}{(b) Warp to canonical coordinate}
\hspace{40mm}{(c) Finetune}

\caption{\textbf{Overview.}
Our method builds on the MLP network $f_\theta$ in NeRF~\cite{Mildenhall-2020-NRS}, and requires only a single image as input at the test time. To learn the geometry and shape priors for single-image view synthesis, we (a) pretrain the model parameter $\theta_p$ using the light stage dataset (\secref{dataset}) consisting of multiple training views and subjects indexed by $m$ (\secref{retraining}). To improve the generalization, we (b) warp the world coordinate $\vec{x}$ to the canonical coordinate derived from the 3D model through a rigid transform $(s,\mathbf{R},\mathbf{t})$, and feed forward the warped coordinate and viewing direction $\vec{d}$ to $f_{\theta}$ (\secref{canonical}) during the view synthesis process. In test time, (c) we finetune the model parameter $\theta_s$ against the input view, and use $f_{\theta_s}$ and volume rendering from color and occupancy ($RGB\sigma$) for view synthesis (\secref{testing}).
}
\label{fig:overview}
\vspace{-2mm}
\end{figure*}
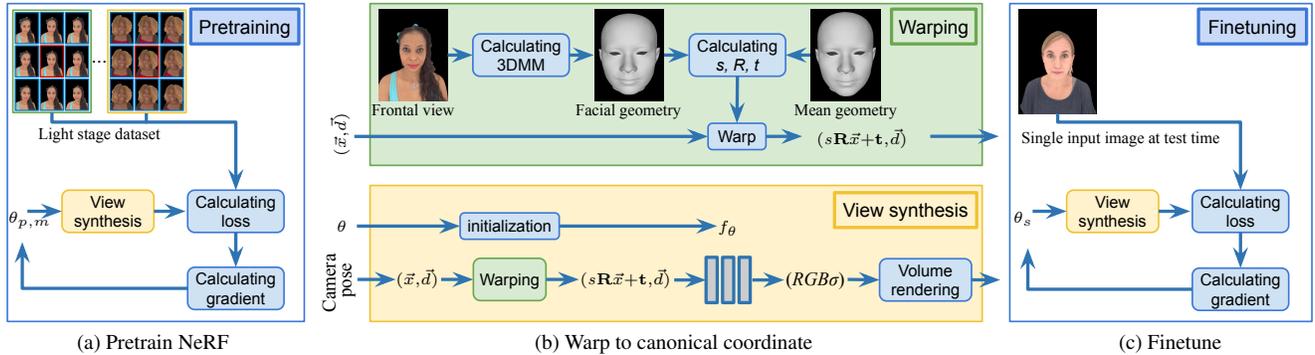

\topic{View synthesis with neural implicit representations.}
Our method builds on recent work of neural implicit representations~\cite{sitzmann2019scene,Mildenhall-2020-NRS,Liu-2020-NSV, Zhang-2020-NAA,Bemana-2020-XIN,Martin-2020-NIT,xian2020space} for view synthesis. 
NeRF~\cite{Mildenhall-2020-NRS} represents the scene as a mapping $\mathcal{F}$ from the world coordinate and viewing direction to the color and occupancy using a compact MLP.
Given a camera pose, one can synthesize the corresponding view by aggregating the radiance over the light ray cast from the camera pose using standard volume rendering.
The MLP is trained by minimizing the reconstruction loss between synthesized views and the corresponding ground truth input images.
Existing methods require tens to hundreds of photos to train a scene-specific NeRF network.
In contrast, our method requires only \emph{one single image} as input. 
We finetune the pretrained weights learned from light stage training data~\cite{Debevec-2000-ATR,Meka-2020-DRT} for unseen inputs.
Compared to the unstructured light field ~\cite{Mildenhall-2019-LLF,Flynn-2019-DVS,Riegler-2020-FVS,Penner-2017-S3R}, volumetric rendering~\cite{Lombardi-2019-NVL,niemeyer2020differentiable}, and image-based rendering~\cite{Hedman-2018-DBF,Hedman-2018-I3P}, our single-image method does not require estimating camera pose~\cite{Schonberger-2016-SFM}, which is challenging at the complex structures and view-dependent properties, like hairs and subtle movement of the subjects between captures.
Similar to the neural volume method~\cite{Lombardi-2019-NVL}, our method improves the rendering quality by sampling the warped coordinate from the world coordinates.

Existing single-image view synthesis methods model the scene with point cloud~\cite{niklaus20193d,Wiles-2020-SEV}, multi-plane image~\cite{Tucker-2020-SVV,huang2020semantic}, or layered depth image~\cite{Shih-CVPR-3Dphoto,Kopf-2020-OS3}. 
Our method focuses on headshot portraits and uses an implicit function as the neural representation. 
We demonstrate foreshortening correction as applications~\cite{Zhao-2019-LPU,Fried-2016-PAM,Nagano-2019-DFN}. 
Instead of training the warping effect between a set of \emph{pre-defined} focal lengths~\cite{Zhao-2019-LPU,Nagano-2019-DFN}, our method achieves the perspective effect at \emph{arbitrary} camera distances and focal lengths.

\topic{Face pose manipulation.}
Conditioned on the input portrait, generative methods learn a face-specific Generative Adversarial Network (GAN)~\cite{Goodfellow-2014-GAN,Karras-2019-ASB,Karras-2020-AAI} to synthesize the target face pose driven by exemplar images~\cite{Wu-2018-RLT,Qian-2019-MAF,Nirkin-2019-FSA,Thies-2016-F2F,Kim-2018-DVP,Zakharov-2019-FSA,wang2019few}, rig-like control over face attributes via face model~\cite{Tewari-2020-SRS,Gecer-2018-SSA,Ghosh-2020-GIF,Kowalski-2020-CCN,sela2017unrestricted,zakharov2020fast}, or learned latent code ~\cite{Deng-2020-DAC,Alharbi-2020-DIG}. 
While these methods yield impressive results and the outputs are photorealistic, these approaches have common artifacts that the generated images often exhibit inconsistent facial features, identity, hairs, and geometries across the results and the input image. 
Our method faithfully preserves the texture and geometry information of the subject across camera poses by using the 3D neural representation invariant to camera poses~\cite{Thies-2019-Deferred,Nguyen-2019-HUL} and taking advantage of pose-supervised training~\cite{Xu-2019-VIG}. 

\topic{3D face modeling.}
Reconstructing face geometry and texture enables view synthesis using graphics rendering pipelines.
Existing single-view methods use the symmetric cues~\cite{Wu-2020-ULP}, morphable model~\cite{Blanz-1999-AMM,Cao-2013-FA3,Booth-2016-A3M,Li-2017-LAM,chai2015high}, mesh template deformation~\cite{Bouaziz-2013-OMF}, regression with deep networks~\cite{Jackson-2017-LP3}, and videos~\cite{Su-2021-ANS}.
However, these \emph{model-based} methods only reconstruct the regions where the model is defined, and do not handle hairs and torsos, or require a separate hair modeling post-processing~\cite{Xu-2020-D3P,Hu-2015-SVH,Liang-2018-VTF}.
Our method combines the benefits from both face-specific modeling and view synthesis on generic scenes. 
We leverage the face domain-specific knowledge by training on a portrait dataset and propose the canonical face coordinates using the 3D face geometric proxy derived by a morphable model.
To model the portrait subject, instead of using facial landmark mesh, we use the finetuned NeRF at the test time to include hairs and torsos.

\topic{Meta-learning.}
Our work is closely related to meta-learning and few-shot learning~\cite{Ravi-2017-OAA,Andrychowicz-2016-LTL,Finn-2017-MAM,chen2019closer,Sun-2019-MTL,Tseng-2020-CDF}. 
To explain the analogy, we consider view synthesis from a camera pose as a \textit{query}, captures associated with the known camera poses from the light stage dataset as \textit{labels}, and training a subject-specific NeRF as a \textit{task}. 
Training NeRFs for different subjects is analogous to training classifiers for various tasks.
At the test time, given a single label from the frontal capture, our goal is to optimize the \textit{testing task}, which learns the NeRF to answer the queries of camera poses.
We leverage gradient-based meta-learning algorithms~\cite{Finn-2017-MAM,Sitzmann-2020-MML} to learn the weight initialization for the MLP in NeRF from the \textit{meta-training tasks}, \ie, learning a single NeRF for different subjects in the light stage dataset.

\topic{Concurrent work on neural volumetric rendering.} 
Several concurrent efforts have been proposed to use neural volumetric rendering for modeling photo-realistic human heads~\cite{raj2021pva,gafni2020dynamic,wang2020learning}.
These methods support synthesizing human heads from a wider range of viewpoints but still require multiple input views.
Recent work also extends NeRF for handling single image input using meta-learning~\cite{tancik2020learned} or pixel-aligned representation~\cite{yu2020pixelnerf}.
Our work focuses on head portrait shots and shows that our use of canonical face coordinates (guided by the fixed 3D morphable model) greatly improves the model generalization to unseen faces. 







\section{Algorithm}
\label{sec:method}


\figref{overview} provides an overview of our method, consisting of pretraining and testing stages.
In the pretraining stage, we train a coordinate-based MLP (same in NeRF) $f_\theta$ on diverse subjects captured from the light stage and obtain the pretrained model parameter optimized for generalization to unseen subjects, denoted as $\theta_p^*$~(\secref{retraining}).
The high diversities among the real-world subjects in identities, facial expressions, and face geometries are challenging for training. 
We address the face shape variation by deforming the world coordinate to the canonical face coordinate using a rigid transform, and train a shared representation across the subjects on the normalized coordinate (\secref{canonical}). 
%
%
At the test time, we initialize the NeRF with our pretrained model parameter $\theta_p^*$ and then finetune it on the frontal view for the input subject $s$. 
We use the finetuned model parameter (denoted by $\theta_s^*$) for view synthesis (\secref{testing}).

%

\subsection{Training data}
\label{sec:dataset}

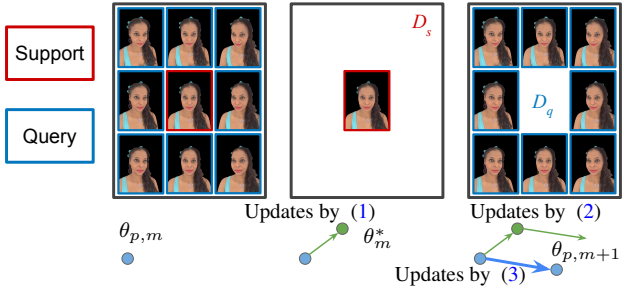
\begin{figure}[t]
\footnotesize
\begin{overpic}[width=1\linewidth]{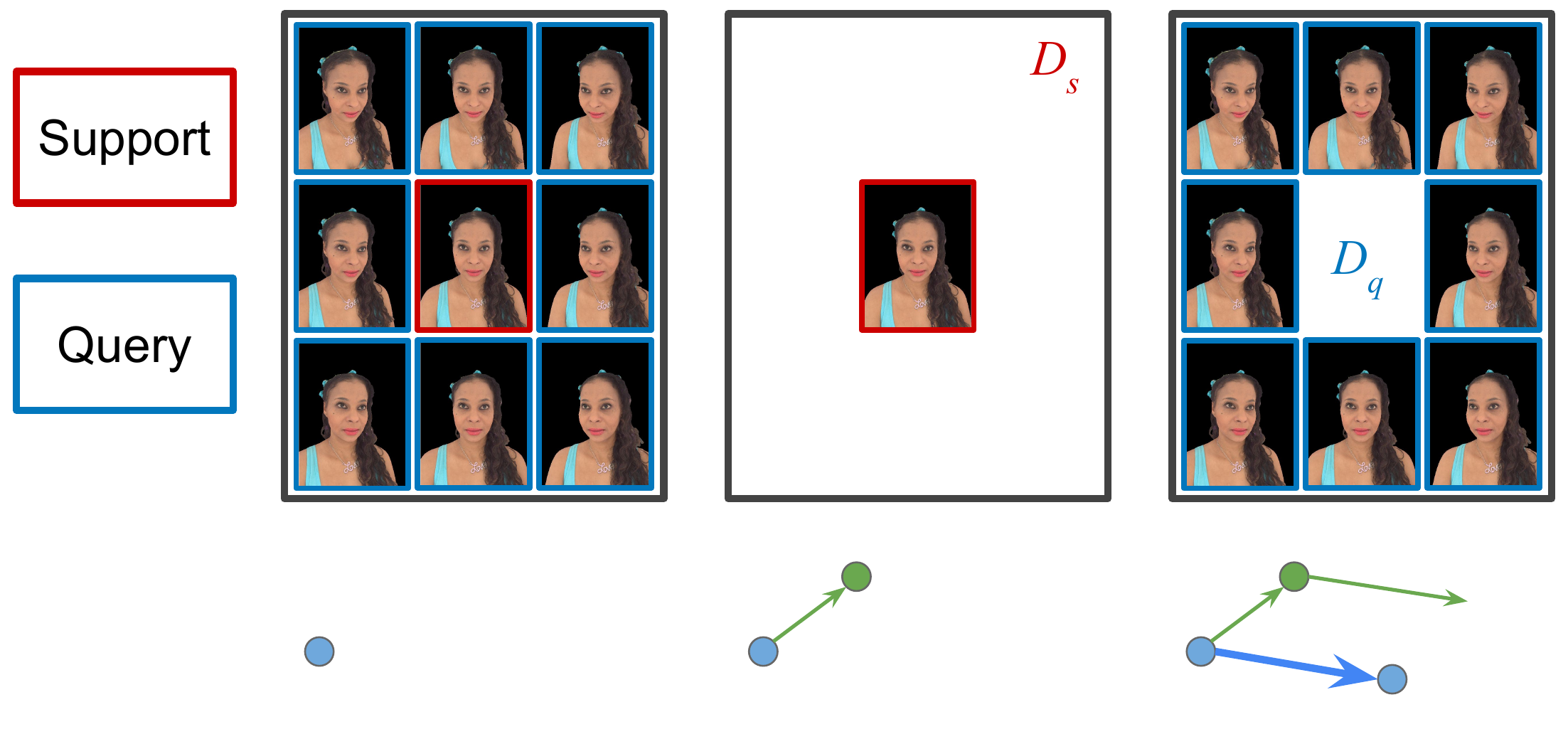}
\put (19,9) {$\theta_{p,m}$}
\put (39,12) {Updates by ~\eqref{eq:inner}}
\put (58,8) {$\theta_{m}^*$}
\put (75,12) {Updates by ~\eqref{eq:update-init-inner-1}}
\put (63,2){Updates by ~\eqref{eq:update-init-inner}}
\put (88,6) {$\theta_{p,m+1}$}

\end{overpic}


\caption{
\textbf{Pretraining with meta-learning framework.} 
The training views of each subject $m$ consists of the frontal view, called support set $D_s$, and the rest views, called query set $D_q$. 
In each iteration, we update the pretrained parameter from $\theta_{p,m}$ to $\theta_{p,m+1}$ using two steps. We first update the parameter to $\theta_m^*$ by finetuning on the $D_s$ using ~\eqref{eq:inner}. Then we continue to update $\theta_m^*$ using~\eqref{eq:update-init-inner-1} on $D_q$, and feedback the update to the pretrained parameter $\theta_{p,m+1}$ using ~\eqref{eq:update-init-inner}. 
%
%
%
%
%
}
\label{fig:pretraining}
\vspace{-2mm}
\end{figure}

Our training data consists of light stage captures over multiple subjects. 
Each subject is lit uniformly under controlled lighting conditions.  
We process the raw data to reconstruct the depth, 3D mesh, UV texture map, photometric normals, UV glossy map, and visibility map for the subject~\cite{Zhang-2020-NLT,Meka-2020-DRT}.
We render a sequence of 5-by-5 training views for each subject by uniformly sampling the camera locations over a solid angle centered at the subject's face at a fixed distance between the camera and subject. 
We span the solid angle by 25\degree~field-of-view vertically and 15\degree~ horizontally.
We set the camera viewing directions to look straight to the subject. 
\figref{pretraining} and supplemental materials show examples of 3-by-3 training views. 
The center view corresponds to the front view expected at the test time, referred to as the \textit{support} set $\mathcal{D}_s$, and the remaining views are the target for view synthesis, referred to as the \textit{query} set $\mathcal{D}_q$.
We refer to the process training a NeRF model parameter for subject $m$ from the support set as a \textit{task}, denoted by $\mathcal{T}_m$. 

%
%
%
%
%
%

%

\subsection{Pretraing NeRF}
\label{sec:retraining}
%

%
%

%
%
%
%
%
%

Our goal is to pretrain a NeRF model parameter $\theta_p^*$ that can easily adapt to capturing an \emph{unseen} subject's appearance and geometry. 
We loop through $K$ subjects in the dataset, indexed by $m=\{0,...,K-1\}$, and denote the model parameter pretrained on the subject $m$ as $\theta_{p,m}$.  
We sequentially train on subjects in the dataset and update the pretrained model as $\{\theta_{p,0}, \theta_{p,1}...,\theta_{p,K-1}\}$, where the last parameter is outputted as the final pretrained model,~\ie, $\theta_p^* = \theta_{p,K-1}$. 
For each task $\mathcal{T}_m$, we train the model on $\mathcal{D}_s$ and $\mathcal{D}_q$ alternatively in an inner loop, as illustrated in ~\figref{pretraining}. 
Since $\mathcal{D}_s$ is available at the test time, we only need to propagate the gradients learned from $\mathcal{D}_q$ to the pretrained model $\theta_p^*$, which transfers the common representations unseen from the front view $\mathcal{D}_s$ alone, such as the priors on head geometry and occlusion. 

\topic{Pretraining on $\mathcal{D}_s$.}
For the subject $m$ in the training data, we initialize the model parameter from the pretrained parameter learned in the previous subject $\theta_{p,m-1}$, and set $\theta_{p,-1}$ to random weights for the first subject in the training loop.
We train a model $\theta_m^*$ optimized for the front view of subject $m$ using the $L_2$ loss between the front view predicted by $f_{\theta_m}$ and $\mathcal{D}_{s}$ 
, denoted as $\mathcal{L}_{\mathcal{D}_s}(f_{\theta_m})$. The optimization iteratively updates the $\theta_m^t$ for $N_s$ iterations as the following:
\begin{align}
\theta_m^{t+1} = \theta_m^t - \alpha \nabla_{\theta} 
\mathcal{L}_{\mathcal{D}_s}(f_{\theta_m^t}),
\label{eq:inner}
\end{align}
where $\theta_m^0 = \theta_{p,m-1}$, $\theta_m^* = \theta_m^{N_s-1}$, and $\alpha$ is the learning rate.

\topic{Pretraining on $\mathcal{D}_q$.}
We proceed to update the pretrained parameter $\theta_{p,m}$ using the loss between the prediction from the known camera pose and the query dataset unseen in test time, as described in the following:
\begin{align}
\theta_m^{t+1} = \theta_m^t - \beta \nabla_{\theta}
\mathcal{L}_{\mathcal{D}_q}(f_{\theta_m^t}), \label{eq:update-init-inner-1} \\
\theta_{p,m}^{t+1} = \theta_{p,m}^t - \beta \nabla_{\theta}
\mathcal{L}_{\mathcal{D}_q}(f_{\theta_m^t}),
\label{eq:update-init-inner}
\end{align}
where $\theta_m^0 = \theta_m^*$ learned from $D_{s}$ in~\eqref{eq:inner}, $\theta_{p,m}^0 =\theta_{p,m-1}$ from the pretrained model on the previous subject, and $\beta$ is the learning rate.
After $N_q$ iterations, we update the pretrained parameter by the following:
\begin{align}
\theta_{p,m} = \theta_{p,m}^{N_q-1}
\label{eq:update-init-outer}
\end{align}
Note that~\eqref{eq:update-init-inner} does not affect the update of the current subject $m$, \ie,~\eqref{eq:update-init-inner-1}, but the gradients are propagated to the next training subject in the subsequent iterations through the pretrained model parameter update in~\eqref{eq:update-init-outer}. 
The training is terminated after visiting the entire dataset over $K$ subjects.
The pseudo-code of the algorithm is described in the supplemental material. 

\topic{Discussion.} 
We assume that the order of applying the gradients learned from $\mathcal{D}_q$ and $\mathcal{D}_s$ are interchangeable, similarly to the first-order approximation in MAML algorithm~\cite{Finn-2017-MAM}. We transfer the gradients from $\mathcal{D}_q$ independently of $\mathcal{D}_s$. 
For better generalization to unseen subjects, the gradients of $\mathcal{D}_s$ will be adapted from the input subject at the test time by finetuning, instead of transferred from the training data.
In our experiments, applying the meta-learning algorithm designed for image classification~\cite{Tseng-2020-CDF} performs poorly for view synthesis.
This is because each update in view synthesis requires gradients gathered from \emph{millions of samples} across the scene coordinates and viewing directions, which do not fit into a single batch in modern GPU.
Our method takes a lot more steps in a single meta-training task for better convergence. 

\subsection{Canonical face space}
\label{sec:canonical}
\newlength\figwidthM
\setlength\figwidthM{0.245\linewidth}

\begin{figure}[t]

\centering%
\parbox[t]{\figwidthM}{\centering%
\includegraphics[width=\figwidthM]{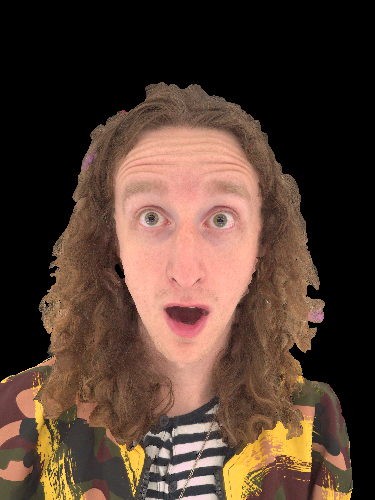}
 \footnotesize (a) Subject}%
\hfill%
\parbox[t]{\figwidthM}{\centering%
\includegraphics[width=\figwidthM]{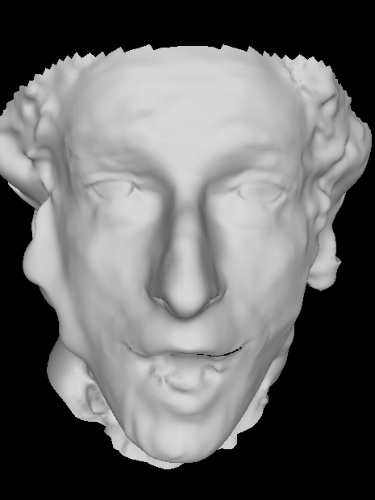}
 \footnotesize (b) World coordinate}%
\hfill%
\parbox[t]{\figwidthM}{\centering%
\includegraphics[width=\figwidthM]{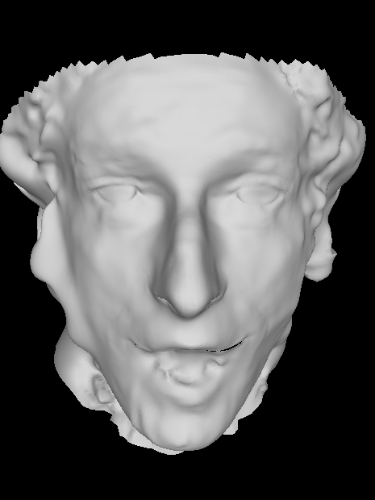}
 \footnotesize (c) Canonical face coordinate}%
\hfill%
\parbox[t]{\figwidthM}{\centering%
\includegraphics[width=\figwidthM]{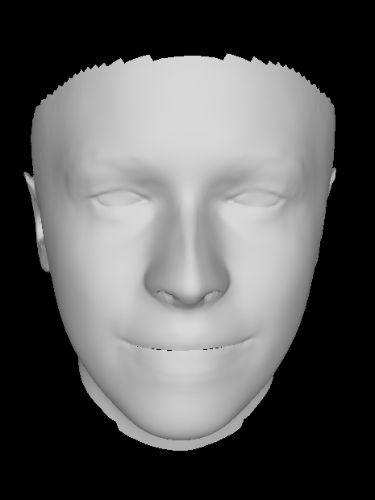}
 \footnotesize (d) Mean face}%



\vspace{1mm}
\caption{
\tb{Rigid transform between the world and canonical face coordinate.} For every subject (a), we detect the face mesh (b), and fit a rigid transform (c) using the vertices correspondence between (b) and the mean face of the dataset (d). We warp from the world (b) to the canonical (c) coordinate, and feed the warped coordinate to NeRF to predict color and occupancy. The meshes (b-d) are for visualization and not used in view synthesis. Only the rigid transform is used in our work.
}
\vspace{-2mm}
\label{fig:3dmm}
\end{figure}
\begin{figure*}[t]
\newlength\figwidthC
\setlength\figwidthC{0.14\linewidth}
\centering%
\parbox[t]{\figwidthC}{\centering%
\includegraphics[trim=30 30 30 80, clip=true, width=\figwidthC]{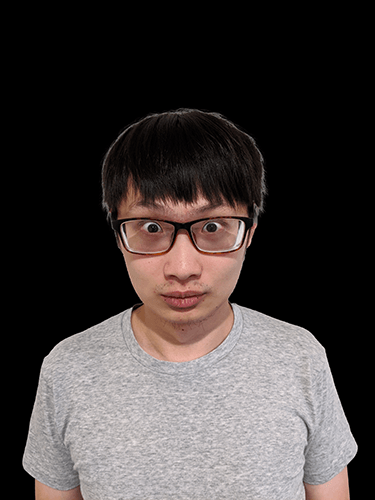}\\%
\includegraphics[trim=40 90 20 20, clip=true, width=\figwidthC]{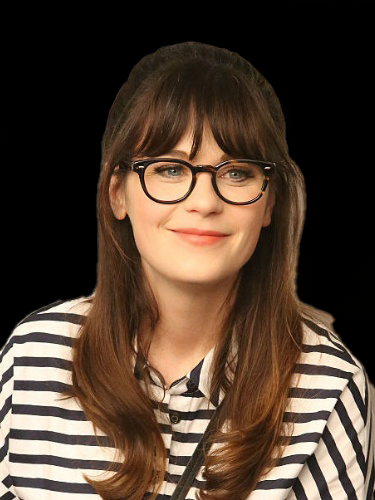}\\%
\includegraphics[trim=0  85 20 70, clip=true, width=\figwidthC]{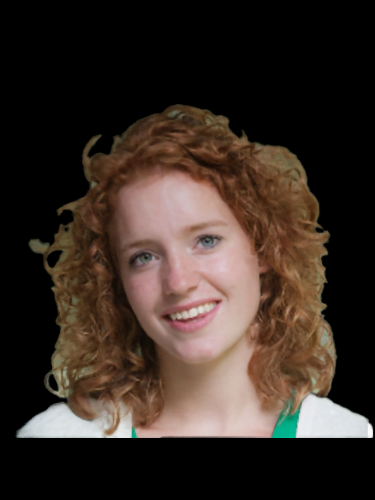}\\%
 \footnotesize (a) Input (front view)}%
\hfill%
\parbox[t]{\figwidthC}{\centering%
\includegraphics[trim=30 30 30 80, clip=true, width=\figwidthC]{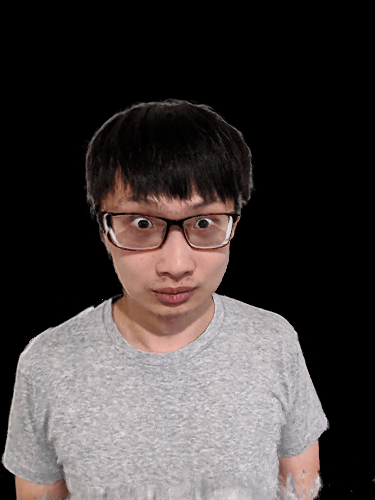}\\%
\includegraphics[trim=40 90 20 20, clip=true, width=\figwidthC]{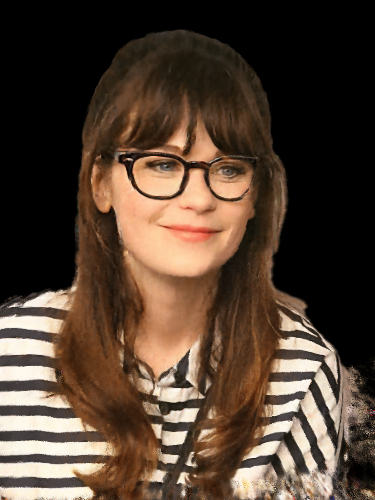}\\%
\includegraphics[trim=0  85 20 70, clip=true, width=\figwidthC]{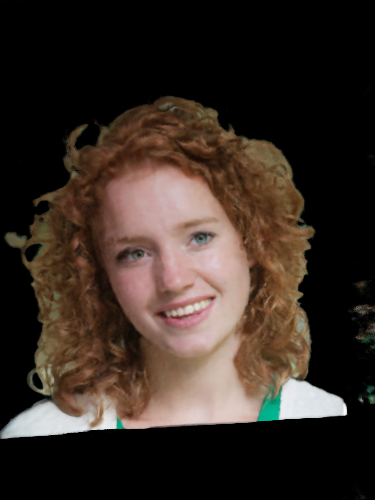}\\%
 \footnotesize (b) Top-left}%
\hfill%
\parbox[t]{\figwidthC}{\centering%
\includegraphics[trim=30 30 30 80, clip=true, width=\figwidthC]{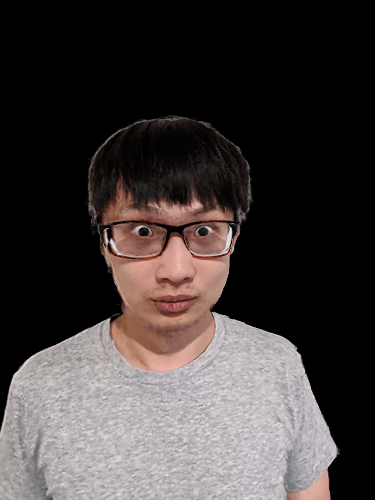}\\%
\includegraphics[trim=40 90 20 20, clip=true, width=\figwidthC]{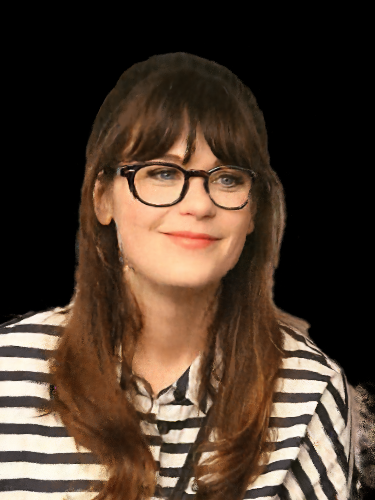}\\%
\includegraphics[trim=0  85 20 70, clip=true, width=\figwidthC]{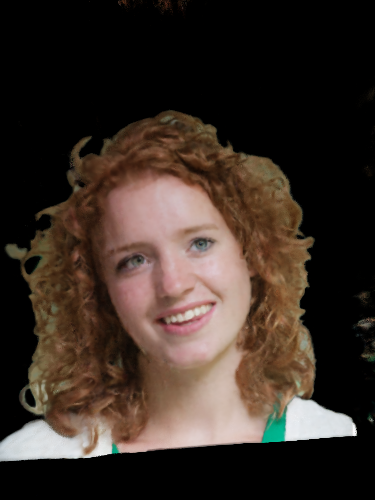}\\%
 \footnotesize (c) Bottom-left}%
\hfill%
\parbox[t]{\figwidthC}{\centering%
\includegraphics[trim=30 30 30 80, clip=true, width=\figwidthC]{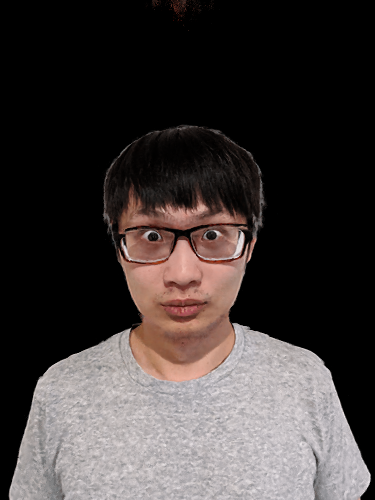}\\%
\includegraphics[trim=40 90 20 20, clip=true, width=\figwidthC]{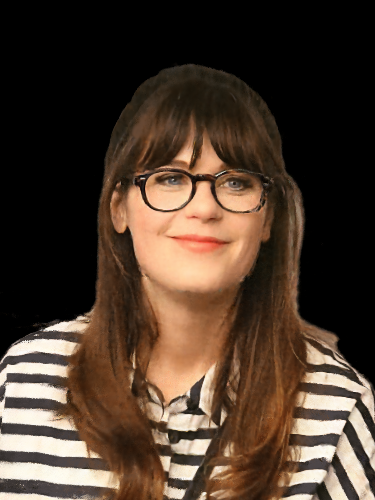}\\%
\includegraphics[trim=0  85 20 70, clip=true, width=\figwidthC]{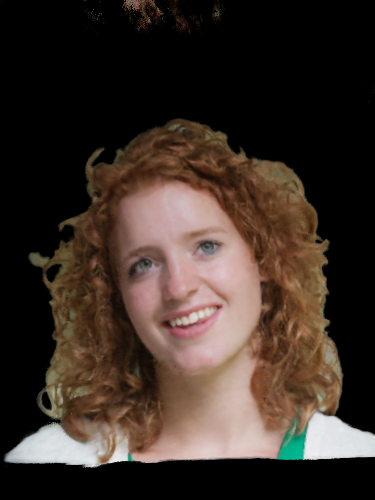}\\%
 \footnotesize (d) Bottom}%
\hfill%
\parbox[t]{\figwidthC}{\centering%
\includegraphics[trim=30 30 30 80, clip=true, width=\figwidthC]{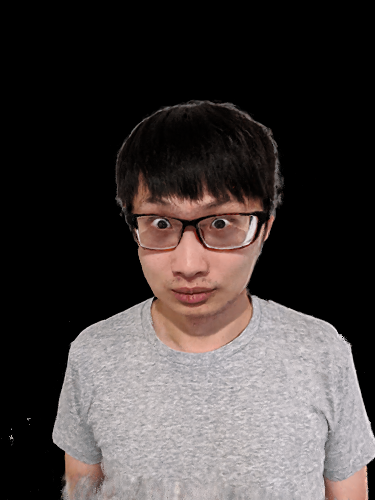}\\%
\includegraphics[trim=40 90 20 20, clip=true, width=\figwidthC]{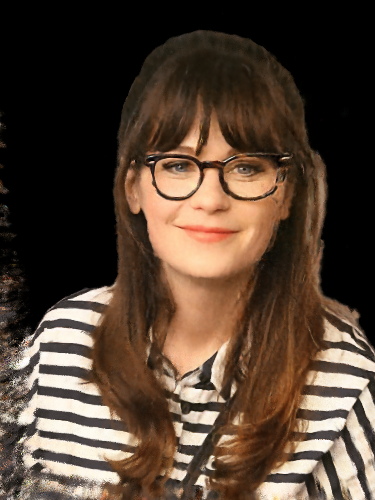}\\%
\includegraphics[trim=0  85 20 70, clip=true, width=\figwidthC]{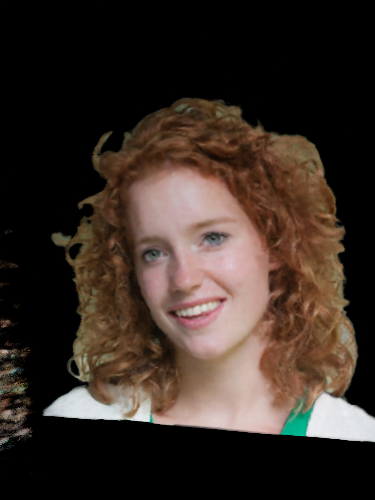}\\%
 \footnotesize (e) Top-Right}%
\hfill%
\parbox[t]{\figwidthC}{\centering%
\includegraphics[trim=30 30 30 80, clip=true, width=\figwidthC]{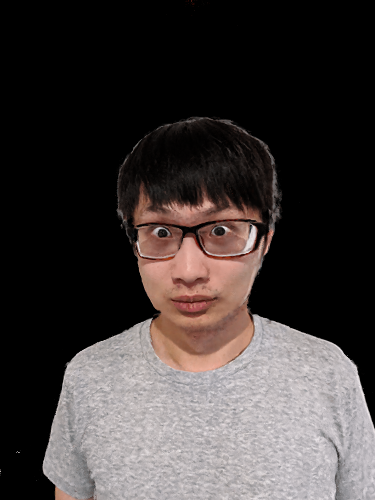}\\%
\includegraphics[trim=40 90 20 20, clip=true, width=\figwidthC]{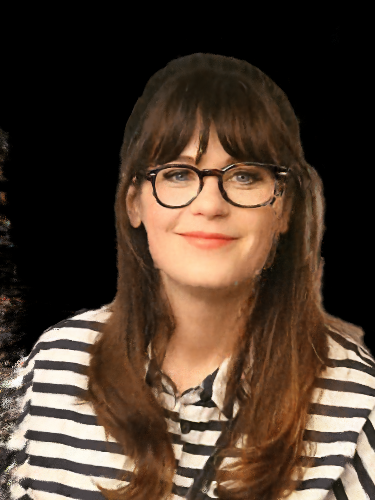}\\%
\includegraphics[trim=0  85 20 70, clip=true, width=\figwidthC]{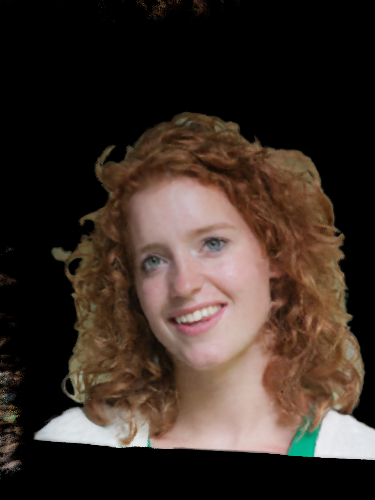}\\%
 \footnotesize (f) Right}%
\hfill 
\parbox[t]{\figwidthC}{\centering%
\includegraphics[trim=30 30 30 80, clip=true, width=\figwidthC]{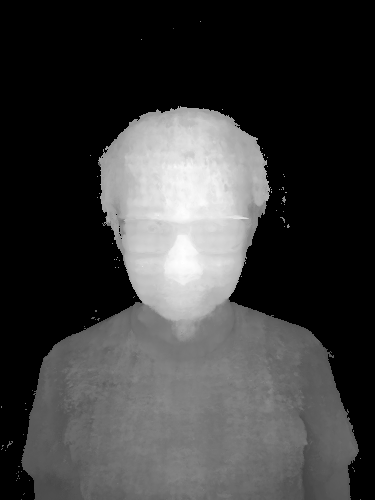}\\%
\includegraphics[trim=40 90 20 20, clip=true, width=\figwidthC]{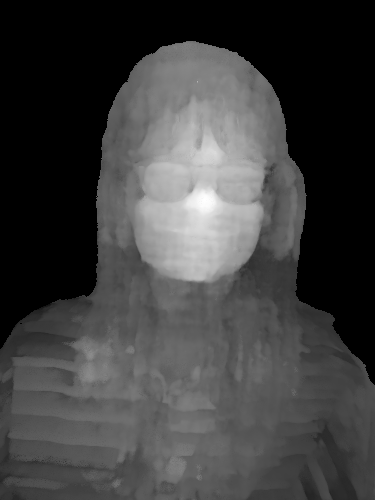}\\%
\includegraphics[trim=0  85 20 70, clip=true, width=\figwidthC]{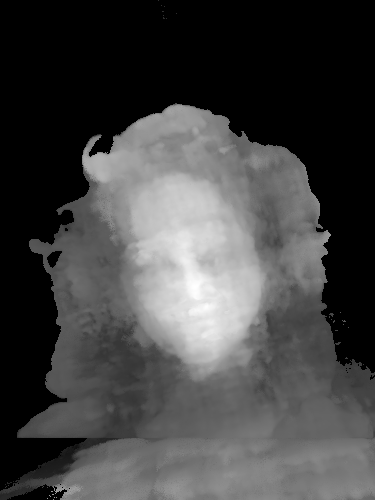}\\%
 \footnotesize (g) Disparity on (a)}%
%


\vspace{1mm}
\caption{\textbf{View synthesis from a single front view}. Our method finetunes the pretrained model on (a), and synthesizes the new views using the controlled camera poses (c-g) relative to (a). The results in (c-g) look realistic and natural. To validate the face geometry learned in the finetuned model, we render the (g) disparity map for the front view (a).
The videos are accompanied in the supplementary materials.
}
\vspace{-1mm}
\label{fig:ours}
\end{figure*}

To address the face shape variations between the training dataset and real-world inputs, we normalize the world coordinate to the canonical space using a rigid transform and apply $f_{\theta}$ on the warped coordinate.
Specifically, for each subject $m$ in the training data, we compute an approximate facial geometry $F_m$ from the frontal image using a 3D morphable model and image-based landmark fitting~\cite{Cao-2013-FA3}.
%
%
We average all the facial geometries in the dataset to obtain the mean geometry $\overline{F}$.
%
%
During the training, we use the vertex correspondences between $F_{m}$ and $\overline{F}$ to optimize a rigid transform by the SVD decomposition (details in the supplemental documents). 
The transform is used to map a point $\mathbf{x}$ in the subject's world coordinate to $\mathbf{x'}$ in the face canonical space: $\mathbf{x'}=s_m \mathbf{R}_m\mathbf{x}+\mathbf{t}_m$, where $s_m, \mathbf{R}_m$ and $\mathbf{t}_m$ are the optimized scale, rotation, and translation minimizing the $L_2$ loss between vertices of $F_{m}$ and $\overline{F}$.
%

During the prediction, we first warp the input coordinate from the world coordinate to the face canonical space through $(s_m,\mathbf{R}_m, \mathbf{t}_m)$. 
We then feed the warped coordinate to the MLP network $f_{\theta}$ to retrieve color and occlusion (\figref{3dmm}). 
We keep the viewing direction $\vec{d}$ in the world coordinate.
The warp makes our method robust to the variation in face geometry and poses in the training and testing inputs, as shown in~\tabref{ablation_warping} and~\figref{ablation_warping}. 
In our method, the 3D model is used to obtain the rigid transform $(s_m,\mathbf{R}_m, \mathbf{t}_m)$. 
We do \emph{not} require the mesh details and priors as in other model-based face view synthesis~\cite{Xu-2020-D3P,Cao-2013-FA3}.



\subsection{Finetuning and rendering}
\label{sec:testing}
%
At the test time, only a single frontal view of the subject $s$ is available.
We first compute the rigid transform described in~\secref{canonical} to map between the world and canonical coordinate. 
Then, we finetune the pretrained model parameter $\theta_p^*$ by repeating the iteration in~\eqref{eq:inner} for the input subject, and output the optimized model parameter $\theta_s^*$.
To render novel views, we sample the camera ray in the 3D space, warp to the canonical space, and feed to $f_{\theta_s^*}$ to retrieve the radiance and occlusion for volume rendering.

\section{Experimental Results}
\label{sec:results}


\newlength\figwidthD
\setlength\figwidthD{0.195\linewidth}

\newlength\figwidthE
\setlength\figwidthE{0.39\linewidth}

\begin{figure}[t]
\footnotesize
\includegraphics[width=1\linewidth]{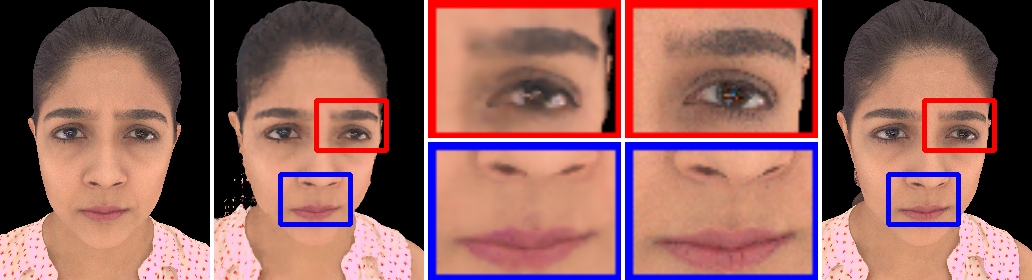}\hfill

\vspace{-4mm}
$\underbracket[1pt][2.0mm]{\hspace{\figwidthD}}_%
    {\substack{\vspace{-2.0mm}\\\colorbox{white}
    {Input}}}$\vspace{1mm} \hfill
$\underbracket[1pt][2.0mm]{\hspace{\figwidthE}}_%
    {\substack{\vspace{-2.0mm}\\\colorbox{white}
    {Our method}}}$\vspace{1mm} \hfill
$\underbracket[1pt][2.0mm]{\hspace{\figwidthE}}_%
    {\substack{\vspace{-2.0mm}\\\colorbox{white}
    {Ground truth}}}$\vspace{1mm}
\caption{\textbf{Comparisons to the ground truth.}
Our results faithfully preserve the details like skin textures, personal identity, and facial expressions from the input.
}
\label{fig:closeup}
\end{figure}

\begin{figure*}[t]
\scriptsize
\newlength\figwidthholo
\setlength\figwidthholo{0.1215\linewidth}
\centering%
\parbox[t]{\figwidthholo}{\centering%
\includegraphics[width=\figwidthholo]{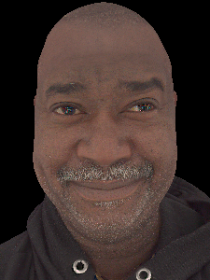}\\%
\includegraphics[width=\figwidthholo]{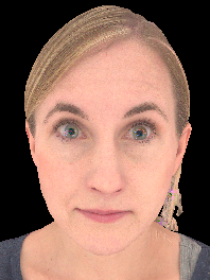}\\%
\includegraphics[width=\figwidthholo]{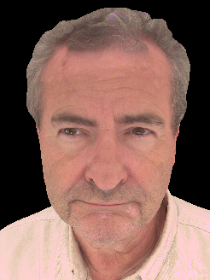}\\%
 \scriptsize Input}%
\hfill%
\parbox[t]{\figwidthholo}{\centering%
\includegraphics[width=\figwidthholo]{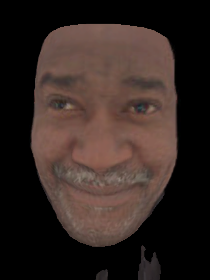}\\%
\includegraphics[width=\figwidthholo]{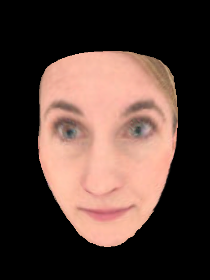}\\%
\includegraphics[width=\figwidthholo]{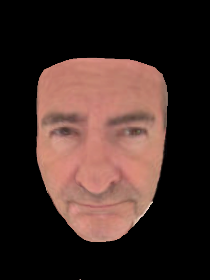}\\%
\scriptsize \mbox{Jackson~\etal \cite{Jackson-2017-LP3}}}%
\hfill%
\parbox[t]{\figwidthholo}{\centering%
\includegraphics[width=\figwidthholo]{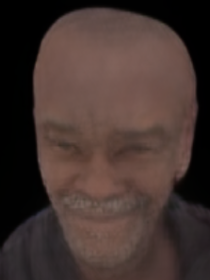}\\%
\includegraphics[width=\figwidthholo]{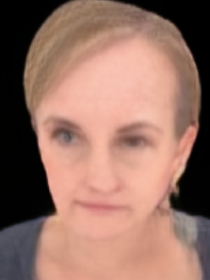}\\%
\includegraphics[width=\figwidthholo]{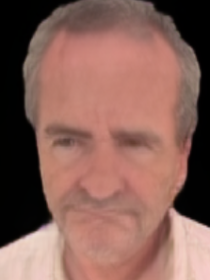}\\%
 \scriptsize\mbox{Siarohin~\etal\cite{siarohin2020first}}}%
\hfill%
\parbox[t]{\figwidthholo}{\centering%
\includegraphics[width=\figwidthholo]{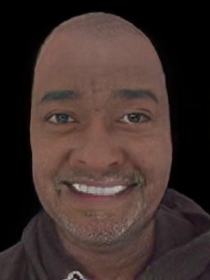}\\%
\includegraphics[width=\figwidthholo]{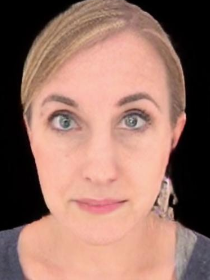}\\%
\includegraphics[width=\figwidthholo]{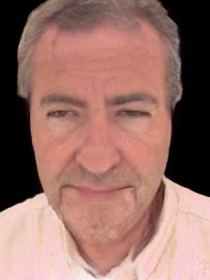}\\%
 \scriptsize Wang~\etal\cite{wang2019few}}%
\hfill%
\parbox[t]{\figwidthholo}{\centering%
\includegraphics[width=\figwidthholo]{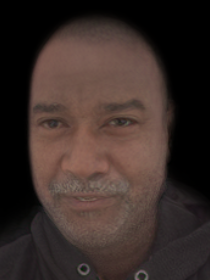}\\%
\includegraphics[width=\figwidthholo]{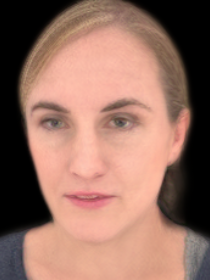}\\%
\includegraphics[width=\figwidthholo]{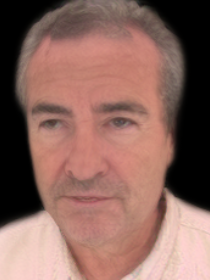}\\%
 \scriptsize Zakharov~\etal\cite{zakharov2020fast}}%
\hfill%
\hfill%
\parbox[t]{\figwidthholo}{\centering%
\includegraphics[width=\figwidthholo]{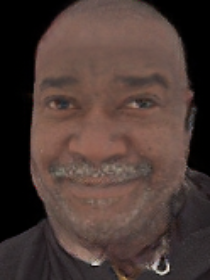}\\%
\includegraphics[width=\figwidthholo]{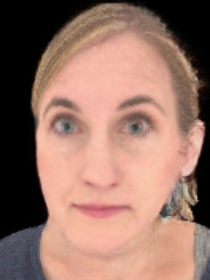}\\%
\includegraphics[width=\figwidthholo]{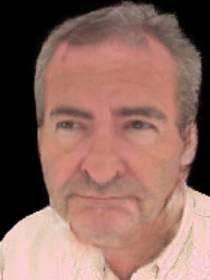}\\%
 \scriptsize Xu~\etal~\cite{Xu-2020-D3P}}%
\hfill%
\parbox[t]{\figwidthholo}{\centering%
\includegraphics[width=\figwidthholo]{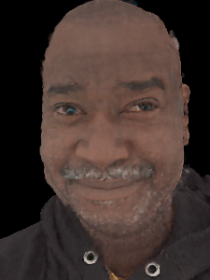}\\%
\includegraphics[width=\figwidthholo]{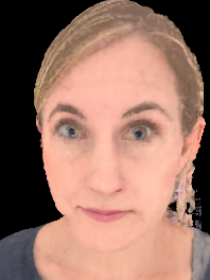}\\%
\includegraphics[width=\figwidthholo]{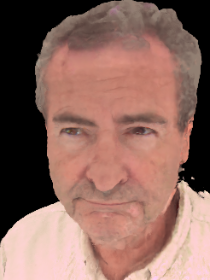}\\%
 \scriptsize Our method}%
\hfill%
\parbox[t]{\figwidthholo}{\centering%
\includegraphics[width=\figwidthholo]{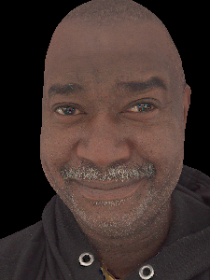}\\%
\includegraphics[width=\figwidthholo]{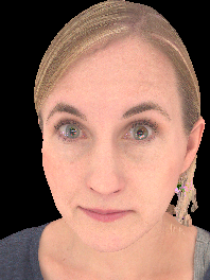}\\%
\includegraphics[width=\figwidthholo]{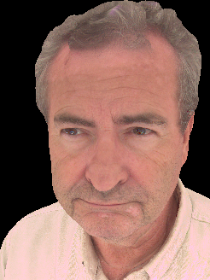}\\%
 \scriptsize Ground truth}%
\vspace{1mm}
\caption{\textbf{Comparison to the state-of-the-art portrait view synthesis on the light stage dataset.} 
In each row, we show the input frontal view and two synthesized views using \cite{Jackson-2017-LP3,Xu-2020-D3P} and our method.
The method by Jackson~\etal~\cite{Jackson-2017-LP3} recovers the geometry only in the center of the face. 
The deep 3D portrait \cite{Xu-2020-D3P} is capable of reconstructing the full head (including hair and ear), but fails to preserve eye gaze, expression, face shape, identity, and hairstyle when comparing to the ground truth. 
The face reenactment methods \cite{siarohin2020first,wang2019few,zakharov2020fast} do not take the camera poses as input, thus a different person with the same ground truth pose to drive the synthesis. The results either lose identity or generate incorrect gaze direction.
Our method produces photorealistic results closest to the ground truth, as shown in the numerical evaluations in Table ~\ref{tab:results}.
}
\label{fig:holodeck_results}
\end{figure*}

Our dataset consists of 70 different individuals with diverse gender, races, ages, skin colors, hairstyles, accessories, and costumes.
For each subject, 
we capture 2-10 different expressions, poses, and accessories on a light stage under fixed lighting conditions.
We include challenging cases where subjects wear glasses.
In total, our dataset consists of 230 captures.
We hold out six captures for testing.
We render the support $\mathcal{D}_s$ and query $\mathcal{D}_q$ by setting the camera field-of-view to 84\degree, a popular setting on commercial phone cameras, and sets the distance to $30$cm to mimic selfies and headshot portraits taken on phone cameras. 
Details about choosing the parameters can be found in the supplemental materials. 


%
%
%

%
\figref{ours} shows our results on the diverse subjects taken in the wild. 
The subjects cover different genders, skin colors, races, hairstyles, and accessories. 
We stress-test the challenging cases like the glasses (the top two rows) and curly hairs (the third row). 
Our results generally look realistic, preserve the facial expressions, geometry, identity from the input, handle the occluded area well, and successfully synthesize the clothes and hairs for the subject.
Our method generalizes well thanks to the finetuning and canonical face coordinate, which closes the gap between the unseen subjects and the pretrained model weights learned from the light stage dataset.
~\figref{ours} also shows that our method is robust when the input face poses are slightly rotated away from the frontal view.
In the supplemental video, we hover the camera in the spiral path to demonstrate the 3D effect.
Our method preserves temporal coherence in challenging areas like hairs and occlusion, such as the nose and ears.

\figref{closeup} compares our results to the ground truth using the subject in the test hold-out set. 
Our method allows precise control of the camera pose and faithfully reconstructs the details from the subject, as shown in the insets. 


%

%
%
%
%


%
%
\begin{table}
\footnotesize
\caption{
\textbf{View synthesis metrics.}
We report the average PSNR, SSIM and LPIPS metrics against the existing methods over the testing set from our light stage dataset.
%
}
\vspace{1mm}
\label{tab:results}
\centering
{
\begin{tabular}{l ccc}
\toprule
&PSNR $\uparrow$ & SSIM $\uparrow$ & LPIPS $\downarrow$\\
	\midrule
Jackson~\etal~\cite{Jackson-2017-LP3}   & 10.23 & 0.4356 & 0.485 \\
Wang~\etal~\cite{wang2019few}           & 14.70 & 0.4578 & 0.380 \\
Zakharov~\etal~\cite{zakharov2020fast}  & 15.25 & 0.4746 & 0.403 \\
Siarohin~\etal~\cite{siarohin2020first} & 15.90 & 0.5149 & 0.437 \\
Xu~\etal~\cite{Xu-2020-D3P}             & 18.91 & 0.5609 & 0.276 \\
Our method                              & \textbf{23.92} & \textbf{0.7688} & \textbf{0.161} \\
\bottomrule
\end{tabular}
\vspace{-2mm}
}
\end{table}
\topic{Comparisons.} \figref{holodeck_results} compares our method to the state-of-the-art face pose manipulation methods~\cite{Xu-2020-D3P,Jackson-2017-LP3} and three face reenactment methods~\cite{siarohin2020first,wang2019few,zakharov2020fast} on six testing subjects held out from the training.
The subjects cover various ages, gender, races, and skin colors.
The results from \cite{Xu-2020-D3P} were kindly provided by the authors.
We obtain the results of the remaining methods using the official implementation with default parameters.
The work by Jackson~\etal~\cite{Jackson-2017-LP3} only covers the face area.
The learning-based head reconstruction method from Xu~\etal~\cite{Xu-2020-D3P} generates plausible results but fails to preserve the gaze direction, facial expressions, face shape, and hairstyles (the bottom row) when comparing to the ground truth.
Since the face reenactment methods do not take the camera poses as input, we use a driving face from a different person with the same ground truth pose for results. 
While these face reenactment methods yield impressive results, they often lose identity or generate incorrect gaze direction.
Our results are objectively closer to the ground truth based on PSNR, and covers the entire subject, including hairs and body.
We report the quantitative evaluation using PSNR, SSIM, and LPIPS~\cite{zhang2018unreasonable} against the ground truth in~\tabref{results}.

\begin{figure}[t]
\footnotesize
\includegraphics[width=0.33\linewidth, trim=0 30 0 30,clip]{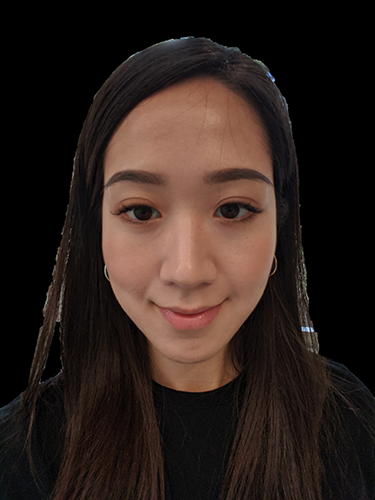}\hfill
\includegraphics[width=0.33\linewidth, trim=0 30 0 30,clip]{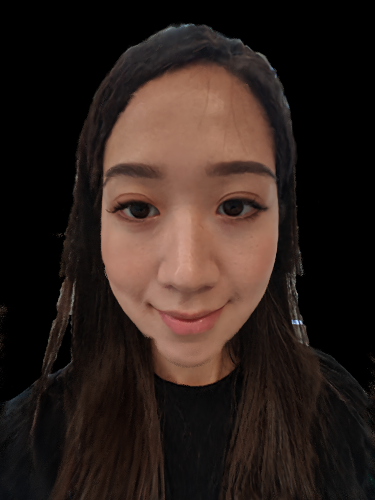}\hfill
\includegraphics[width=0.33\linewidth, trim=8 37 8 41, clip]{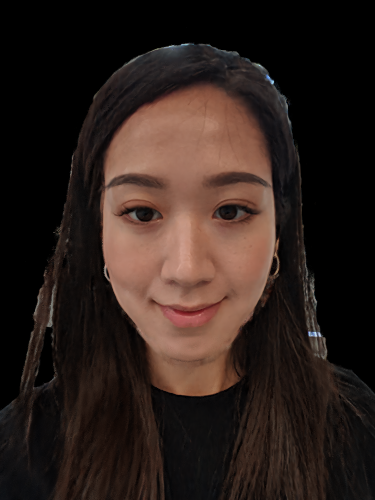}\hfill
\includegraphics[width=0.33\linewidth, trim=0 30 0 40,clip]{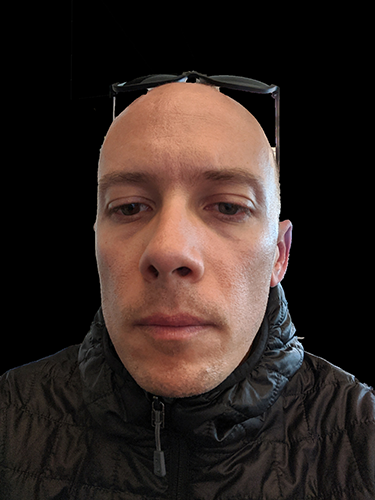}\hfill
\includegraphics[width=0.33\linewidth, trim=0 30 0 40,clip]{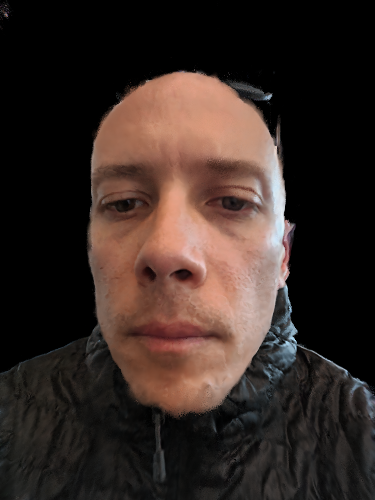}\hfill
\includegraphics[width=0.33\linewidth, trim=8 37 8 51, clip]{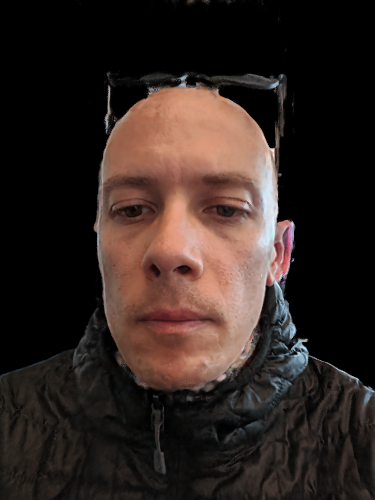}\hfill

\mpage{0.3}{(a) Input} \hfill
\mpage{0.3}{(b) Closer to subject} \hfill
\mpage{0.3}{(c) Further} \hfill

\caption{\textbf{Perspective effect manipulation.}
Given an input (a), we virtually move the camera closer (b) and further (c) to the subject, while adjusting the focal length to match the face size.
We manipulate the perspective effects such as dolly zoom in the supplementary materials.
}
\label{fig:perspective}
\end{figure}

\begin{table}
\footnotesize
    \caption{
    \textbf{Ablation study on initialization methods.}
    }
    \vspace{1mm}
    \label{tab:ablation_initializations}
    \centering
    \begin{tabular}{l | ccc}
    \toprule
    Initialization &PSNR $\uparrow$ & SSIM $\uparrow$ & LPIPS $\downarrow$\\
    \midrule
    Random              & 14.99 & 0.5763 & 0.491 \\
    Pretrain by ~\eqref{eq:min-loss}             & 22.87 & 0.7824 & 0.215 \\
    Our method       & \textbf{23.70} & \textbf{0.8051} & \textbf{0.178} \\
    \bottomrule
    \end{tabular}
\end{table}
\begin{table}
\footnotesize
    \caption{
    \textbf{Ablation study on canonical face coordinate.}
    %
    }
    \vspace{1mm}
    \label{tab:ablation_warping}
    \centering
    {
    \begin{tabular}{l | ccc}
    \toprule
    Coordinate &PSNR $\uparrow$ & SSIM $\uparrow$ & LPIPS $\downarrow$\\
    \midrule
    World  & 24.80 & 0.8167 & 0.172 \\
    Canonical (our method)    & \textbf{24.98} & \textbf{0.8178} & \textbf{0.156} \\
    \bottomrule
    \end{tabular}
    }
\end{table}

\topic{Perspective manipulation.}
Portraits taken by wide-angle cameras exhibit undesired foreshortening distortion due to the perspective projection ~\cite{Fried-2016-PAM,Zhao-2019-LPU}.
As our representation is 3D in nature, we can render a new image by virtually moving the camera closer (or further) from the subject and adjusting the focal length correspondingly to preserve the face area.
We demonstrate perspective effect manipulation using portrait NeRF in~\figref{perspective} and the supplemental video.
When the camera sets a longer focal length, the nose looks smaller, and the portrait looks more natural. 
Since our training views are taken from a single camera distance, the vanilla NeRF rendering~\cite{Mildenhall-2020-NRS} requires inference on the world coordinates outside the training coordinates and leads to the artifacts when the camera is too far or too close, as shown in the supplemental materials. 
We address the artifacts by re-parameterizing the NeRF coordinates to infer on the training coordinates.


\begin{figure}[t]
\footnotesize
\includegraphics[width=0.245\linewidth]{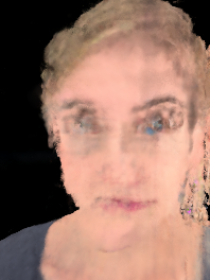}\label{fig:random-init}\hfill
\includegraphics[width=0.245\linewidth]{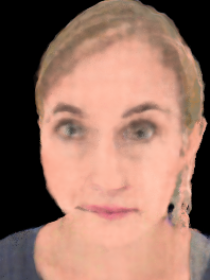}\label{fig:min-loss-init}\hfill
\includegraphics[width=0.245\linewidth]{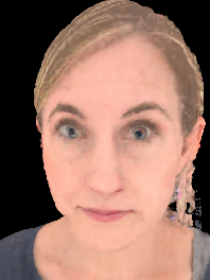}\label{fig:our-init}\hfill
\includegraphics[width=0.245\linewidth]{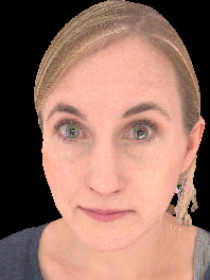}\hfill

\mpage{0.22}{(a) Random} \hfill
\mpage{0.22}{\mbox{(b) Pretrain by \eqref{eq:min-loss}}} \hfill
\mpage{0.22}{(c) Our method} \hfill
\mpage{0.22}{(d) GT} \hfill
\vspace{1mm}
\caption{
\textbf{Ablation study on different weight initialization.}
(a) Training the MLP from scratch using random weight initialization fails to converge and leads to poor results. 
(b) Pretraining using ~\eqref{eq:min-loss} suffers from blurry rendering and artifacts.
(c) Our method produces the best quality compared to the ground truth (d).
\label{fig:ablation_initializations}
}
\end{figure}
\begin{figure}[t]
\footnotesize
\includegraphics[trim=0 50 0 110, clip=true, width=0.328\linewidth]{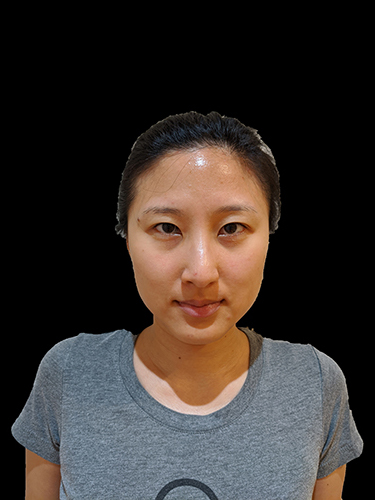}
\includegraphics[trim=0 50 0 110, clip=true, width=0.328\linewidth]{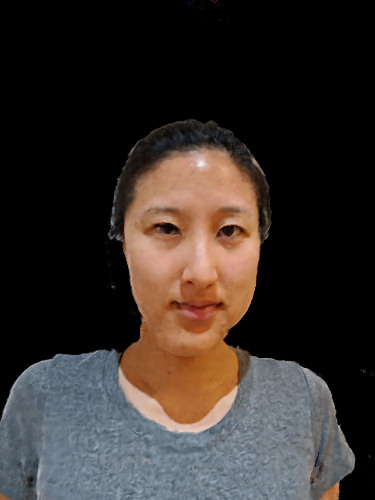}~\label{fig:world-coordinate}
\includegraphics[trim=0 50 0 110, clip=true, width=0.328\linewidth]{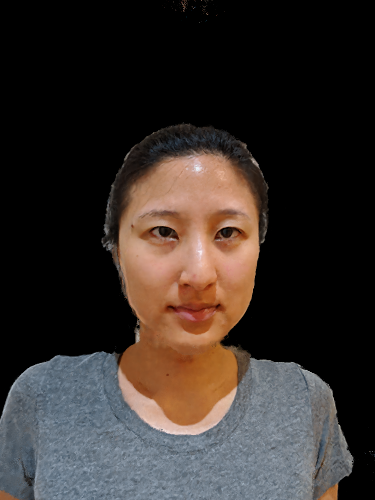}~\label{fig:our-coordinate}

\mpage{0.3}{(a) Input} \hfill
\mpage{0.3}{(b) World coordinate} \hfill
\mpage{0.3}{(c) Our method} \hfill
\caption{\textbf{Ablation study on face canonical coordinates.} Our method using (c) canonical face coordinate shows better quality than using (b) world coordinate on chin and eyes.
\label{fig:ablation_warping}
}
\end{figure}
\begin{figure}[t]
\footnotesize
\includegraphics[width=0.245\linewidth]{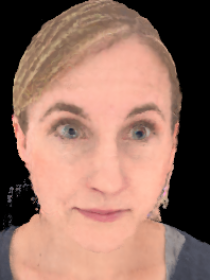}\hfill
\includegraphics[width=0.245\linewidth]{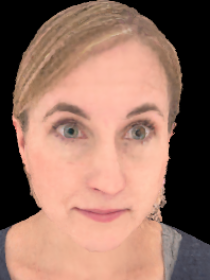}\hfill
\includegraphics[width=0.245\linewidth]{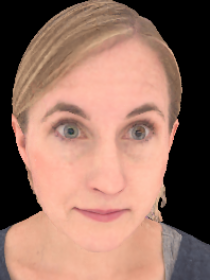}\hfill
\includegraphics[width=0.245\linewidth]{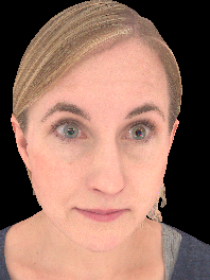}\hfill
\mpage{0.22}{(a) 1 view \\ PSNR = 24.98} \hfill
\mpage{0.22}{(b) 2 views \\ PSNR = 27.70} \hfill
\mpage{0.22}{(c) 5 views \\ PSNR = 32.45} \hfill
\mpage{0.22}{(d) GT\\~} \hfill
\vspace{1mm}
\caption{\textbf{Ablation study on the number of input views during testing.}
Our results improve when more views are available.
} 
\label{fig:ablation_split}
\end{figure}

\begin{table}
\footnotesize
    \caption{
    \textbf{Ablation study on training sizes.}
    %
    }
    \vspace{1mm}
    \label{tab:ablation_training_tasks}
    \centering
    {
    \begin{tabular}{c | ccc}
    \toprule
    Training size &PSNR $\uparrow$ & SSIM $\uparrow$ & LPIPS $\downarrow$\\
    \midrule
    18 & 24.61 & 0.8110 & 0.163 \\
    27 & 24.98 & \textbf{0.8178} & \textbf{0.156} \\
    59 & \textbf{25.04} & 0.8129 & 0.164 \\
    100 & 24.65 & 0.8120 & 0.165 \\
    \bottomrule
    \end{tabular}
    }
\end{table}
\begin{table*}[t]
\footnotesize
    \caption{
    \textbf{Ablation study on number of input views.}
    %
    }
    \vspace{1mm}
    \label{tab:ablation_split}
    \centering
    {
    \begin{tabular}{c  ccc c ccc c ccc}
    \toprule
    & \multicolumn{3}{c}{Random initialization} && \multicolumn{3}{c}{Pretrain with ~\eqref{eq:min-loss}} && \multicolumn{3}{c}{Our method}\\
    \cline{2-4} \cline{6-8} \cline{10-12}
    Number of input views & PSNR $\uparrow$ & SSIM $\uparrow$ & LPIPS $\downarrow$ && PSNR $\uparrow$ & SSIM $\uparrow$ & LPIPS $\downarrow$ && PSNR $\uparrow$ & SSIM $\uparrow$ & LPIPS $\downarrow$\\
    \midrule
    1  & 17.34 & 0.6988 & 0.347 && 24.47 & 0.8109 & 0.175 && \textbf{24.98} & \textbf{0.8178} & \textbf{0.156} \\
    2  & 23.21 & 0.7936 & 0.214 && 27.37 & 0.8445 & 0.130 && \textbf{27.70} & \textbf{0.8647} & \textbf{0.115} \\
    5  & 31.12 & 0.8958 & 0.094 && 31.57 & 0.8826 & 0.107 && \textbf{32.45} & \textbf{0.9045} & \textbf{0.090} \\
    \bottomrule
    \end{tabular}
    }
\end{table*}

\subsection{Ablation study}
\heading{Initialization.}
\figref{ablation_initializations} compares the results finetuned from different initialization methods.
Without any pretrained prior, the random initialization~\cite{Mildenhall-2020-NRS} in~\figref{ablation_initializations}(a) fails to learn the geometry from a single image and leads to poor view synthesis quality. 
Next, we pretrain the model parameter by minimizing the $L_2$ loss between the prediction and the training views across all the subjects in the dataset as the following:
\begin{align}
\theta^* = \argmin_\theta \sum_m
\mathcal{L}_{\mathcal{D}_s}(f_{\theta}) + \mathcal{L}_{\mathcal{D}_q}(f_{\theta}),  
\label{eq:min-loss}
\end{align}
where $m$ indexes the subject in the dataset. 
\figref{ablation_initializations}(b) shows that such a pretraining approach can also learn geometry prior from the dataset but shows artifacts in view synthesis. 
The synthesized face looks blurry and misses facial details.
Our pretraining in~\figref{ablation_initializations}(c) outputs the best results against the ground truth. 
The quantitative evaluations are shown in~\tabref{ablation_initializations}.  

\heading{Canonical face coordinate.}
\figref{ablation_warping} and~\tabref{ablation_warping} compare the view synthesis using the face canonical coordinate (\secref{canonical}) to the world coordinate. 
Without warping to the canonical face coordinate, the results using the world coordinate in~\figref{ablation_warping}(b) show artifacts on the eyes and chins. 
Our method outputs a more natural look on face in~\figref{ablation_warping}(c), and performs better on quality metrics against ground truth across the testing subjects, as shown in~\tabref{ablation_warping}.


\heading{Training task size.}
Our method does not require a large number of training tasks consisting of many subjects. 
In~\tabref{ablation_training_tasks}, we show that the validation performance saturates after visiting 59 training tasks. 
To balance the training size and visual quality, we use 27 subjects for the results shown in this paper. 
Compared to the majority of deep learning face synthesis works, \eg,~\cite{Xu-2020-D3P}, which require \emph{thousands of individuals} as the training data, the capability to generalize portrait view synthesis from a smaller subject pool makes our method more practical to comply with the privacy requirement on personally identifiable information.


\heading{Input views in test time.}
Our method can incorporate multi-view inputs associated with known camera poses to improve the view synthesis quality. 
At the finetuning stage, we compute the reconstruction loss between each input view and the corresponding prediction.
We show the evaluations on different number of input views against the ground truth in~\figref{ablation_split} and comparisons to different initialization in~\tabref{ablation_split}. 
Compared to the vanilla NeRF using random initialization~\cite{Mildenhall-2020-NRS}, our pretraining method is highly beneficial when very few (1 or 2) inputs are available.
The margin decreases when the number of input views increases and is less significant when 5+ input views are available.  

\subsection{Limitations}
\begin{figure}
\footnotesize
\includegraphics[trim=20 65 20 90, clip=true, width=0.27\linewidth]{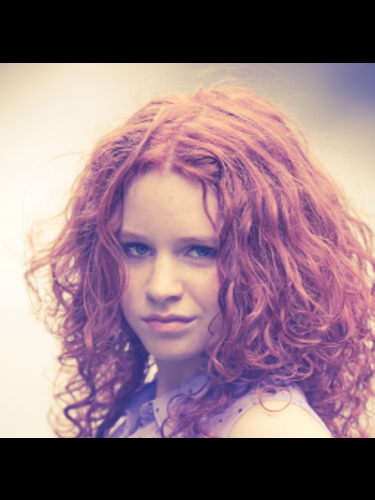}\hfill
\includegraphics[trim=20 65 20 90, clip=true, width=0.27\linewidth]{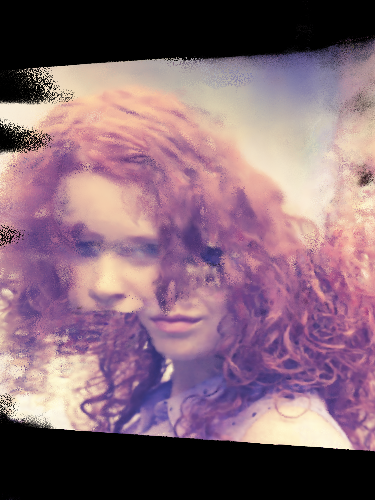}\hfill
\includegraphics[trim=0 35 0 10, clip=true, width=0.23\linewidth]{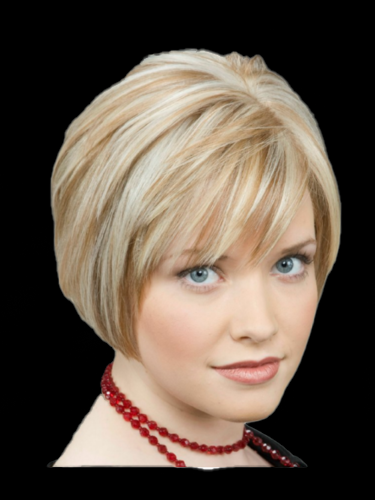}\hfill
\includegraphics[trim=0 35 0 10, clip=true, width=0.23\linewidth]{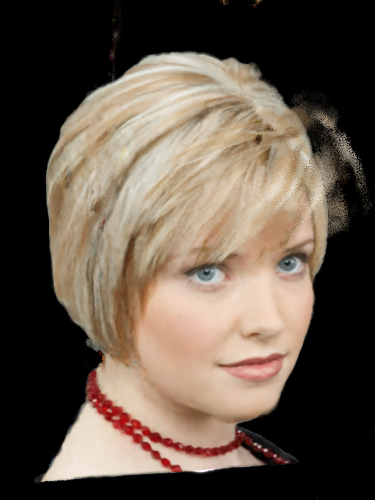}\hfill

\mpage{0.48}{(a) Background} \hfill
\mpage{0.48}{(b) Non-frontal view} \hfill


\caption{\textbf{Limitations.} Left and right in (a) and (b): input and output of our method. (a) When the background is not removed, our method cannot distinguish the background from the foreground and leads to severe artifacts. (b) When the input is not a frontal view, the result shows artifacts on the hairs.
}
\label{fig:limitation-profile}
\end{figure}

While our work enables training NeRF from a single portrait image, there are several limitations.
First, our method cannot handle the background, which is diverse and difficult to collect on the light stage (\figref{limitation-profile}(a)). 
Users can use off-the-shelf person segmentation~\cite{Wadhwa-2018-SDW} to separate the foreground, inpaint the background~\cite{Liu-2018-IIF}, and composite the synthesized views to address the limitation. 
Second, our method requires the input subject to be roughly in frontal view and does not work well with the profile view (\figref{limitation-profile}(b)). 
Extrapolating the camera pose to the unseen poses from the training data is challenging and could lead to artifacts.
Third, as we build on the volume rendering in NeRF, our method is similarly slow at the runtime.
We believe that the slow rendering speed can be alleviated by integrating with recent advances on speeding up volumeric rendering~\cite{lindell2020autoint,liu2020neural,neff2021donerf}.
Fourth, our work focuses only on a static image and leaves the extension to dynamic portrait video to future work.

\vspace{-2.5mm}
\section{Conclusions}
\label{sec:conclusions}

We presented a method for portrait view synthesis using a single headshot photo. 
Our method builds upon the recent advances of neural implicit representation and addresses the limitation of generalizing to an unseen subject when only one single image is available.
Specifically, we leverage gradient-based meta-learning for pretraining a NeRF model so that it can quickly adapt using light stage captures as our meta-training dataset.
We also address the shape variations among subjects by learning the NeRF model in canonical face space. 
We validate the design choices via ablation study and show that our method enables natural portrait view synthesis compared with state of the arts.

{\small
\bibliographystyle{ieee_fullname}
\bibliography{main}
}

\end{document}